\documentclass[conference]{IEEEtran}
\IEEEoverridecommandlockouts
\usepackage{cite}
\usepackage{amsmath,amssymb,amsfonts}
\usepackage{algorithmic}
\usepackage{graphicx}
\usepackage{textcomp}
\usepackage{xcolor}
\usepackage{bbding}
\usepackage{booktabs}
\usepackage{multirow}
\usepackage[hidelinks]{hyperref}
\usepackage{bm}
\usepackage{balance}
\usepackage[T1]{fontenc}

\usepackage{marvosym}

\def\BibTeX{{\rm B\kern-.05em{\sc i\kern-.025em b}\kern-.08em
    T\kern-.1667em\lower.7ex\hbox{E}\kern-.125emX}}

\def \eg {\emph{e.g.}}
\def \ie {\emph{i.e.}}

\begin{document}

\title{GCA-SUNet: A Gated Context-Aware Swin-UNet for Exemplar-Free Counting}

\author{

\IEEEauthorblockN{
Yuzhe Wu$^{1,}$\IEEEauthorrefmark{1}, Yipeng Xu$^{1,}$\IEEEauthorrefmark{1}, 
Tianyu Xu$^{1}$, 
Jialu Zhang$^{1}$, 
Jianfeng Ren$^{1,2,}$\textsuperscript{\Letter}, Xudong Jiang$^{3}$
}
\IEEEauthorblockA{
$^{1}$ \text{School of Computer Science}, University of Nottingham Ningbo China}
\IEEEauthorblockA{
$^{2}$ \text{Nottingham Ningbo China Beacons of Excellence Research and Innovation Institute}, \\ University of Nottingham Ningbo China}
\IEEEauthorblockA{
$^{3}$ \text{School of Electrical \& Electronic Engineering}, \text{Nanyang Technological University}, Singapore}

\tt\small{
\{scyyw16,ssyyx20,scytx2,jialu.zhang,jianfeng.ren\}@nottingham.edu.cn,exdjiang@ntu.edu.sg}

\thanks{\IEEEauthorrefmark{1}Equally contribution. \textsuperscript{\Letter}Corresponding author.}
\thanks{This work was supported by the Ningbo Science and Technology Bureau under Grants 2022Z173, 2022Z217, 2023Z138, 2023Z237 and 2024Z110.}
}

\maketitle

\begin{abstract}
Exemplar-Free Counting aims to count objects of interest without intensive annotations of objects or exemplars. To achieve this, we propose a Gated Context-Aware Swin-UNet (GCA-SUNet) to directly map an input image to the density map of countable objects. Specifically, a set of Swin transformers form an encoder to derive a robust feature representation, and a Gated Context-Aware Modulation block is designed to suppress irrelevant objects or background through a gate mechanism and exploit the attentive support of objects of interest through a self-similarity matrix. The gate strategy is also incorporated into the bottleneck network and the decoder of the Swin-UNet to highlight the features most relevant to objects of interest. By explicitly exploiting the attentive support among countable objects and eliminating irrelevant features through the gate mechanisms, the proposed GCA-SUNet focuses on and counts objects of interest without relying on predefined categories or exemplars. Experimental results on the real-world datasets such as FSC-147 and CARPK demonstrate that GCA-SUNet significantly and consistently outperforms state-of-the-art methods. The code is available at \href{https://github.com/Amordia/GCA-SUNet}{https://github.com/Amordia/GCA-SUNet}.
\end{abstract}


\section{Introduction}
\label{sec:intro}
Object counting determines the number of instances of a specific object class in an image,
\eg, vehicles~\cite{Wentao_2025_PR, Jialu_2024_AAAI}, crowd~\cite{dai2023crowd}, and cells~\cite{xie2018microscopy}. 
It can be broadly categorized into the following. 
1)~Class-Specific Counting (CSC), counting specific categories like fruits~\cite{bellocchio2019fruit} and animals~\cite{barbedo2020counting};
2)~Class-Agnostic Counting (CAC), counting objects based on visual exemplars~\cite{ranjan2021learning, shi2022bmnetplus, liu2023countr} or text prompts~\cite{xu2023zero, kang2023vlcounter}; 
3)~Exemplar-Free Counting (EFC), counting objects without exemplars, presenting a significant challenge in discerning countable objects and determining their repetitions~\cite{ranjan2022exemplar, hobley2022learning, liu2023countr}.

Exemplar-Free Counting shows promise for automated systems such as wildlife monitoring~\cite{akccay2020automated}, healthcare~\cite{zavrtanik2020segmentation}, and anomaly detection~\cite{liu2023simplenet}.
Hobley and Prisacariu directly regressed the image-level features learned by attention modules into a density map~\cite{hobley2022learning}.  
CounTR~\cite{liu2023countr} and LOCA~\cite{djukic2023loca} are originally designed for CAC tasks, but can be adapted to EFC tasks by using trainable components to simulate exemplars. 
RepRPN-Counter identifies exemplars from region proposals by majority voting~\cite{ranjan2022exemplar}, and DAVE selects valuable objects using a strategy similar to majority vote~\cite{pelhan2024dave}.

Existing models~\cite{liu2023countr, djukic2023loca, pelhan2024dave} often explicitly require exemplars to count similar objects. EFC methods such as RepRPN-Counter do not require exemplars but generate them through region proposal~\cite{ranjan2022exemplar}. Either explicit or implicit exemplars may induce sample bias as exemplars can't cover the sample distribution. 
To address the challenge, we propose Gated Context-Aware Swin-UNet (GCA-SUNet), which directly maps an input image to the density map of countable objects, without any exemplars. 
Specifically, the encoder consists of a set of Swin Transformers to extract features, and a set of Gated Context-Aware Modulation (GCAM) blocks to exploit the attentive supports of countable objects. 
The bottleneck network includes a Gated Enhanced Feature Selector (GEFS) to emphasize the encoded features that are relevant to countable objects. 
The decoder includes a set of Swin transformers for generating the density map, 
with the help of Gated Adaptive Fusion Units (GAFUs) to selectively weigh features based on their relevance to countable objects.  
Finally, a regression head is utilized to derive the density map from the aggregated features.

One key challenge in EFC is to effectively differentiate countable objects from other objects.
The GCAM blocks tackle this challenge by first evaluating the feature quality by computing the feature score for each token, and then prioritizing those with informative content. To uncover the informative features, GCAM computes pairwise similarities between tokens through a self-similarity matrix, exploiting the support of repeating objects in the same scene. Lastly, a gate mechanism is incorporated to highlight the most relevant features while suppressing irrelevant ones.  

Another challenge is that foreground objects often share similar low-level features with background content. The skip connections directly fuse low-level features in the encoder with high-level semantics in the decoder, potentially impeding counting performance as the background could disturb the foreground objects. To tackle this issue, gate mechanisms are incorporated into both GEFS and GAFU to suppress irrelevant low-level features while preserving as much information on objects of interest as possible. The former selectively enhances the condensed features at the bottleneck, and the latter filters the irrelevant features in the decoder. 

Our contributions can be summarized as follows.
1)~The proposed GCA-SUNet achieves exemplar-free counting through a UNet-like architecture that utilizes Swin transformer blocks for feature encoding and decoding, avoiding the sample bias of exemplar-based methods~\cite{ranjan2022exemplar}. 
2)~The proposed GCAM focuses on countable objects by exploiting attentive support of repetitive objects through the self-similarity matrix. 
3)~The gate mechanism is integrated into various modules, \eg, GCAM, GEFS and GAFU, which suppresses the features of irrelevant objects or background while highlighting the most relevant features to countable objects. 
4)~Our GCA-SUNet is evaluated on the FSC-147 and CARPK datasets. It outperforms state-of-the-art methods for exemplar-free counting. 

\section{Related Work}
Object counting has evolved through various paradigms, each with obvious tradeoffs on annotation overhead, adaptability, and generalization. Existing methods can be broadly grouped into Class-Specific Counting (CSC), Class-Agnostic Counting (CAC) and Exemplar-Free Counting (EFC).

\noindent\textbf{Class-Specific Counting.} Early object counting efforts are tightly coupled with class-specific detectors trained on predefined categories~\cite{hsieh2017drone, li2019coda, qiu2019crowd}. These frameworks offer accurate localization and counting, but their reliance on class-specific training data hinders their ability to generalize to unseen object classes. As a result, scaling these approaches to more diverse scenarios remains challenging and computationally expensive. 

\noindent\textbf{Class-Agnostic Counting.} To overcome the limitations of class specificity, CAC methods~\cite{ranjan2021learning, shi2022bmnetplus, liu2023countr} have been developed. Instead of modelling each class individually, these methods learn a universal mapping from image features to density maps, guided by visual exemplars~\cite{ranjan2021learning, shi2022bmnetplus, liu2023countr, zhang2023counting} or textual prompts~\cite{xu2023zero, kang2023vlcounter}. This strategy alleviates the need for category-specific annotations and enables broader generalization. Prominent examples include Generic Matching Network (GMN)~\cite{lu2019class}, which infers counts through similarity comparisons between query images and exemplar shots, and FamNet~\cite{ranjan2021learning}, which refines query features through feature correlation and adaptive parameter updates. CounTR~\cite{liu2023countr} enhances the token of exemplars and the query image using cross-attention mechanics. DAVE~\cite{pelhan2024dave} utilizes a detector to generate a high-recall object set and excludes outliers by cosine-similarity metrics. Models exploiting textual exemplars~\cite{xu2023zero, kang2023vlcounter} often rely on rich language models and vision language pre-training to interpret instructions and align them with visual content. Despite these advancements, the above methods inherently remain dependent on the quality and relevance of the guidance, which might falter if these cues are not informative or biased.

\noindent\textbf{Exemplar-Free Counting.} In contrast, Exemplar-Free Counting (EFC) eliminates all external guidance, \ie, no predefined categories, no exemplars, and no textual prompts. Instead, EFC models autonomously discover and emphasize regions containing countable objects solely from visual cues~\cite{ranjan2022exemplar, hobley2022learning, liu2023countr}. This reduces annotation burden and prevents biases introduced by external supports, making EFC more suitable for diverse and unpredictable real-world settings. For instance, some approaches leverage pre-trained vision transformers~\cite{ranjan2022exemplar} with weakly-supervised training, or automatically generate region proposals as exemplars ~\cite{hobley2022learning}, facilitating the counting without any provided exemplar. 

While prior models have progressively moved from strict CSC tasks to more flexible EFC tasks, the complete absence of guiding signals beyond the visual images remains an ambitious goal. The proposed Gated Context-Aware Swin-UNet (GCA-SUNet) tackles this challenge by directly predicting a density map for countable objects without any exemplar or textual guidance. More importantly, by strategically integrating gate mechanisms within the encoder, bottleneck, and decoder, and leveraging self-similarity to filter out irrelevant backgrounds, the proposed method advances the state-of-the-art in Exemplar-Free Counting significantly. This strategy achieves both robustness and scalability, opening new possibilities for counting arbitrary objects in unconstrained scenarios.

\section{Proposed Method}

\label{sec:methods}
\subsection{Overview of Proposed Method}

\begin{figure*}[!t]
\centering
\includegraphics[width=0.88\textwidth]{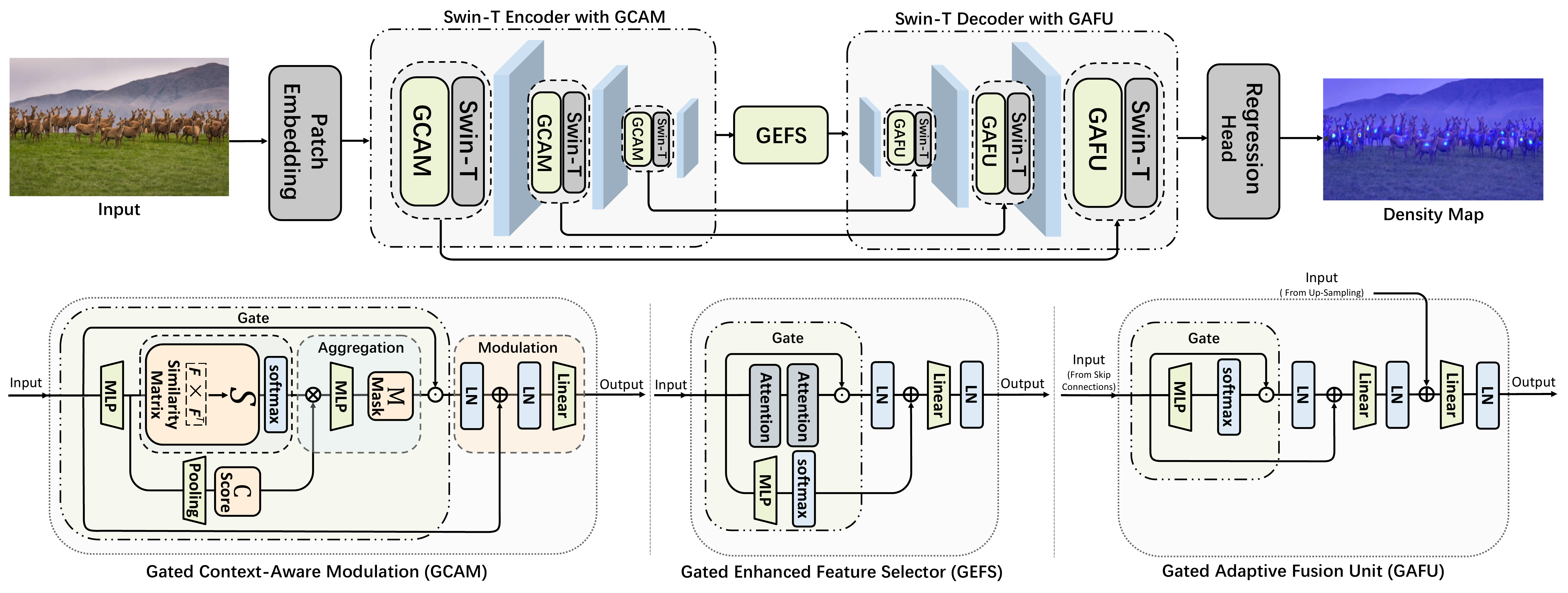}

\caption{Overview of proposed GCA-SUNet. It consists of an encoder, bottleneck, and a decoder. The encoder consists of a set of GCAM blocks to highlight the features relevant to countable objects while suppressing others, and a set of Swin transformers to extract features. The GEFS in the bottleneck and the GAFU in the decoder enhance the features of objects of interest. Finally, a regression head generates a density map for estimating the number of objects.}
\label{fig:arch}
\end{figure*}

The proposed GCA-SUNet is built upon a Swin-UNet architecture~\cite{swinunet}, with three new types of building blocks, GCAM, GEFS and GAFU to exploit attentive support of countable objects and suppress irrelevant tokens or features, as outlined in Fig.~\ref{fig:arch}. 
The primary motivation to choose Swin-UNet is that it has demonstrated superior performance in many tasks~\cite{Xingke_2025_AAAI}. It is also well suited for our task in which we need to generate object density maps of the same size as the input image. 
Specifically, the proposed GCA-SUNet begins with patched image feature $\bm{F}$, following by feature encoding, 
\begin{equation}
\bm{F}^\text{E}_{i} = \mathcal{F}_i^{\text{Down}}(\mathcal{F}_i^{\text{Swin-T}}(\mathcal{F}_i^{\text{GCAM}}(\bm{F}_{i-1}^\text{E}))),
\end{equation}
where $\mathcal{F}_i^{\text{Down}}$, $\mathcal{F}_i^{\text{Swin-T}}$, $\mathcal{F}_i^{\text{GCAM}}$ denote down-sampling, GCAM, and Swin-T processing, and $\bm{F}_{i-1}^\text{E}$ and $\bm{F}_{i}^\text{E}$ are the input and output features at the $i$-th stage, respectively. GCAM enhances the token for countable objects and suppresses others.

At the bottleneck, the features are enhanced through GEFS, \ie, 
$\bm{F}^{\text{BN}} = \mathcal{F}^{\textsc{GEFS}}(\bm{F}_{K}^\text{E})$, 
where $\mathcal{F}^{\textsc{GEFS}}(\cdot)$ denotes the operation of GEFS, and $\bm{F}_{K}^\text{E}$ denotes the output features at the $K$-th stage of the encoder. GEFS selects the features corresponding to the countable object using a gate mechanism. 

Subsequently, a set of Swin transformer blocks are utilized as the decoder to derive the density map. Specifically, the features at the $j$-th stage of the decoder are derived as, 
\begin{equation}
\bm{F}^{\text{D}}_{j} = \mathcal{F}^\text{Up}_j(\mathcal{F}^{\text{Swin-T}}_j(\mathcal{F}^{\text{GAFU}}_j(\bm{F}^{\text{D}}_{j-1}, \bm{F}_{K+1-j}^\text{E}))), 
\end{equation}
where $\mathcal{F}^\text{Up}_j$, $\mathcal{F}^{\text{Swin-T}}_j$, and  $\mathcal{F}^{\text{GAFU}}_j$ denote the operation of up-sampling, Swin transformer, and GAFU block, respectively. The GAFU enhances features through a gate mechanism, prioritizing crucial information with a dynamic assigned weight. 

Finally, features are processed through a regression head, 
\begin{equation}
\bm{F}^{\text{head}} = \mathcal{F}^{\text{Head}}(\bm{F}^{\text{D}}_{K}), 
\end{equation}
where $\mathcal{F}^{\text{Head}}$ denotes the regression head consisting of a series of convolutional blocks. The output is a density map that accurately represents the object count.

\subsection{Swin-T Encoder with GCAM}
A set of SWIN transformer blocks form the encoder to extract features related to countable objects. The GCAM adapts a dynamic token modulation to simultaneously exploit the attentive support of tokens relevant to countable objects and suppress features of irrelevant objects~\cite{Chenglin_2024_ESWA, Jialu_2024_ACMMM, Chengtai_2025_AAAI}.
This process facilitates self-probing among objects and precise capture of objects of the same category for exemplar-free counting.

Specifically, we first compress token features $\bm{F}_{i}^\text{E}$ using an MLP, ${\bm{F}_{i}^\text{proj}} = \mathcal{F}^\textsc{MLP}\left(\bm{F}_{i}^\text{E}\right)$.
To identify the objects of interest, we resort to two key observations: 1) The objects should be salient enough to step out from the background; 2) Similar objects could support each other to boost the saliency. 
The former is exploited by computing the average feature score $\bm{C}_i$ for each token through average pooling $\mathcal{F}^\textsc{AVG}$ as,
\begin{equation}
\bm{C}_i = \mathcal{F}^\textsc{AVG}\left( {\bm{F}_{i}^\text{proj}} \right).   
\end{equation}   
The score reflects the importance of tokens, prioritizing those with rich content. 
Tokens that frequently appear in similar contexts are more likely to be related to the target object of interest. To identify them, we employ a similarity matrix,
\begin{equation}
\bm{S}_i = \sigma \left( {\bm{F}_{i}^\text{proj}}\times {\bm{F}_{i}^\text{proj}}^T \right),
\end{equation}  
where $\sigma$ is a softmax function to normalize similarities across rows.
$\bm{S}_i$ captures the semantic similarity of tokens in a spatial context to emphasize tokens that repeatedly share similar features, thereby emphasizing potential countable objects. 
A mask $\bm{M}_i$ is derived by aggregating  $\bm{S}_i$ and $\bm{C}_i$ as,  
\begin{equation}
\bm{M}_i = \sigma\left(\mathcal{F}^\textsc{MLP}\left( \bm{S}_i \times \bm{C}_i \right) \right).
\end{equation}
$\bm{C}_i$ encodes the token importance when considering the token alone, while $\bm{S}_i$ encodes the token importance after interacting with other tokens. 
The tokens are then filtered by the mask as, 
\begin{equation}
{\bm{F}_{i}^\text{GCAM}} = \mathcal{F}^{\text{Linear}}(\mathcal{F}^{\text{LN}}(\mathcal{F}^{\text{LN}}(\bm{F}_{i}^\text{E} \odot \bm{M}_i)+\bm{F}_{i}^\text{proj}))), 
\end{equation}
where $\odot$, $\mathcal{F}^{\text{LN}}$ and $\mathcal{F}^{\text{Linear}}$ denote element-wise product, layer normalization and linear layer, respectively. The GCAM applies the mask $\bm{M}_i$ to $\bm{F}_{i}^\text{E}$, filtering out less relevant features and reinforcing those critical ones for countable objects.

The proposed GCAM selectively amplifies the importance of tokens related to significant object features through pairwise similarities.  It is significantly different from LOCA~\cite{djukic2023loca} and DAVE~\cite{pelhan2024dave} which depends on predefined prototypes to predict object densities. Instead, our GCAM leverages the self-similarity matrix for more dynamic and precise modulation of features. It is also different from RCC~\cite{hobley2022learning} that relies on global feature comparisons, and CounTR~\cite{liu2023countr} that utilizes attention-driven similarity matrices. The GCAM emphasizes a clear distinction between relevant and irrelevant tokens. 

\begin{table*}[htbp]
    \caption{Comparison with other methods on the FSC-147 dataset~\cite{ranjan2021learning}, with best results highlighted in bold.}%
    \label{table1}
	\centering
    \renewcommand{\arraystretch}{1} 
	\setlength{\tabcolsep}{8pt}
	\begin{tabular}{cllcccccc}
	\toprule
	\multirow{2}[2]{*}{Category} &\multirow{2}[2]{*}{Method} & \multirow{2}[2]{*}{Backbone}& \multirow{2}[2]{*}{Resolution} & \multicolumn{2}{c}{Test Set} & \multicolumn{2}{c}{Val Set}  \\

	\cmidrule(lr){5-6} \cmidrule(lr){7-8}
	& & & & MAE & RMSE & MAE & RMSE \\
    \midrule
    \multirow{5}{*}{CAC}
    & GMN~\cite{lu2019class}~\footnotesize{ACCV'18}& ResNet-50 &255 &37.86 &141.39 & 39.02& 106.06 \\
	 & FamNet~\cite{ranjan2021learning}~\footnotesize{CVPR'21} &ResNet-50 &384 & 32.27 & 131.46 & 32.15 & 98.75  \\
     & LOCA~\cite{djukic2023loca}~\footnotesize{ICCV'23} &ResNet-50&512 & 16.22 & 103.96 & 17.43 & 54.96\\
    & DAVE~\cite{pelhan2024dave}~\footnotesize{CVPR'24}&ResNet-50&512  & 15.14 & 103.49& \textbf{15.54 }&\textbf{52.67} \\
    & CounTR~\cite{liu2023countr}~\footnotesize{BMVC'22}&ViT-B \& CNN& 384& 14.71 & 106.87 &  18.07 & 71.84  \\ 
    \midrule
    \multirow{3}[2]{*}{EFC}
	& RepRPN-C~\cite{ranjan2022exemplar}~\footnotesize{ACCV'22} &ResNet-50&[384,1504]& 26.66 & 129.11 & 29.24 & 98.11  \\
    & RCC~\cite{hobley2022learning}~\footnotesize{CVPRW'23} &ViT-B& 224&17.12 & 104.53 &17.49 & 58.81 \\
    \cmidrule{2-8}
	& \textbf{Proposed GCA-SUNet}& SwinT-B & 384 &\textbf{14.08}&	\textbf{84.77} &  17.02 & 55.37	\\
\bottomrule
	\end{tabular}
	\label{tab:ExpSOTA}
\end{table*}

\subsection{Bottleneck with GEFS} 
The proposed GEFS selectively filters out features in the bottleneck that are semantically irrelevant to the objects of interest, while allowing critical condensed features to pass. It is implemented by first deriving the local token weights as
$\bm{W}^{\text{GEFS}} = \sigma(\mathcal{F}^{\text{MLP}}(\bm{F}_{K}^\text{E}))$,
and applying them on features as, 
\begin{equation}
\bm{F}_{0}^\text{D} = \bm{F}_{K}^\text{E} + (\bm{W}^{\text{GEFS}} \odot (\mathcal{F}^{\textsc{Attn.}}(\mathcal{F}^{\textsc{Attn.}}(\bm{F}_{K}^\text{E})))). 
\end{equation}
The GEFS is positioned at the bottleneck where features transit from the down-sampling pathway to the up-sampling pathway. As a vital bottleneck, the GEFS compresses and filters essential object-related features, ensuring that only the most relevant information of countable objects is advanced into the up-sampling pathway. Specifically, the attention blocks allow the model to capture a more comprehensive range of relationships and refine its ability to extract high-level semantic representations, resulting in more accurate and semantically enriched feature embeddings. Furthermore, a gate mechanism is incorporated into GEFS to selectively prioritize specific aspects of this condensed representation, effectively filtering out less relevant semantics. This process not only refines features by strengthening inter-dependencies between relevant features, but also lays a solid foundation for comprehensive reconstruction of the up-sampling pathway. 

\subsection{Swin-T Decoder with GAFU}
The decoder contains a set of Swin-T blocks to articulate the density map, and a set of Gated Adaptive Fusion Units (GAFUs) to integrate low-level features from the encoder with abstract features from the up-sampling pathway.  
In each GAFU, the token weights are determined by a gate mechanism as, $\bm{W}^{\text{GAFU}} = \sigma(\mathcal{F}^{\text{MLP}}(\bm{F}^{\text{E}}_{i}))$,  
and modulate the features as,
\begin{equation} 
\bm{F}^{\text{G}}_{i} = \bm{F}^{\text{E}}_{i} +  (\bm{W}^{\text{GAFU}} \odot \bm{F}^{\text{E}}_{i}). 
\end{equation} 
Subsequently, these features are fused with the decoder features as  ${\bm{F}^{\text{GAFU}}_{j}} = \mathcal{F}^\text{Linear}([\bm{F}^{\text{D}}_{j}, \bm{F}^{\text{G}}_{i}])$.
By weighing the features during the fusion process, the GAFU effectively concentrates on semantic information pertinent to countable objects, minimizing the interference from irrelevant details.

\subsection{Discussion of Gate Mechanisms}
Gate mechanisms are incorporated into the three proposed modules, GCAM, GEFS and GAFU. 
They share a similar underlying mechanism, buth they are tailored to address distinct challenges. 1)~Gates in GCAM tackle the challenges of differentiating countable objects from others, by emphasizing the relevant tokens through a similarity-based gate. It effectively probes countable objects within the same category through a similarity matrix. 2)~Gates in GEFS or GAFU tackle the challenges of effectively differentiating foreground features from the background context, \ie, to suppress irrelevant low-level features while preserving essential information on objects of interest. By reinforcing relevant inter-dependencies, it prioritizes the semantic information of countable objects, thereby enhancing its ability to distinguish the objects from the background when constructing the density map.

\section{Experimental Results}
\label{sec:exp}

\subsection{Experimental Settings}
Two benchmark datasets are utilized for evaluation. Following~\cite{liu2023countr, djukic2023loca, pelhan2024dave}, we employ the Mean Average Error (MAE) and Root Mean Squared Error (RMSE) as evaluation metrics. 

\noindent \textbf{FSC-147}~\cite{ranjan2021learning} is collected from real world and it consists of 6,135 images of 147 categories with 7 to 3,731 objects per image, mainly composed of foods, animals, kitchen utensils, and vehicles. It is officially split into 3,659, 1,286 and 1,190 images for training, validation and testing, respectively.

\noindent \textbf{CARPK}~\cite{hsieh2017drone} comprises 1,448 images taken from four parking lots using a bird's-eye view. It is primarily intended for object counting and vehicle localization tasks, and it is officially split into 989 training images and 459 testing images.

The Swin-T blocks are pre-trained on ImageNet-22k~\cite{liu2021swin}, and other modules are randomly initialized. AdamW optimizer is employed, with an initial learning rate of 0.003, a decay rate of 0.95 and a batch size of 16. The model is trained with a warm-up period of 50 epochs and 300 epochs in total. The input image size is $384\times384$. Data augmentation~\cite{liu2023countr} is employed to facilitate efficient training. All experiments are conducted using two NVIDIA RTX A5000 GPUs. 

\subsection{Comparison Results on FSC-147 Dataset}
Comparison experiments are conducted on the FSC-147 dataset. 
The results are summarized in
Table~\ref{table1}. Following the practice in~\cite{hobley2022learning, liu2023countr}, we report the errors on the test and validation sets. We have the following observations.
1) The proposed GCA-SUNet outperforms all the compared methods regarding test errors, while performing slightly poorer than DAVE~\cite{pelhan2024dave} regarding validation errors. The superior results demonstrate its effectiveness for EFC tasks. It outperforms not only the dedicated methods for solving EFC tasks such as RepRPN-C~\cite{ranjan2022exemplar} and RCC~\cite{hobley2022learning}, but also state-of-the-art models for CAC tasks. Compared to the second-best method, CounTR~\cite{liu2023countr}, the performance gain is 0.63 for of MAE and 16.47 for RMSE.
2) Although the GCA-SUNet performs slightly poorer than DAVE~\cite{pelhan2024dave} in terms of validation errors, it significantly outperforms DAVE in terms of test errors, \ie, a performance gain of 1.06 for MAE and 18.72 for RMSE. DAVE performs well on the validation set and generalizes poorly on the test set. In contrast, the GCA-SUNet generalizes well on the novel test set with minimal errors. 

\begin{figure}[!t]
\centering
\includegraphics[width=\linewidth]{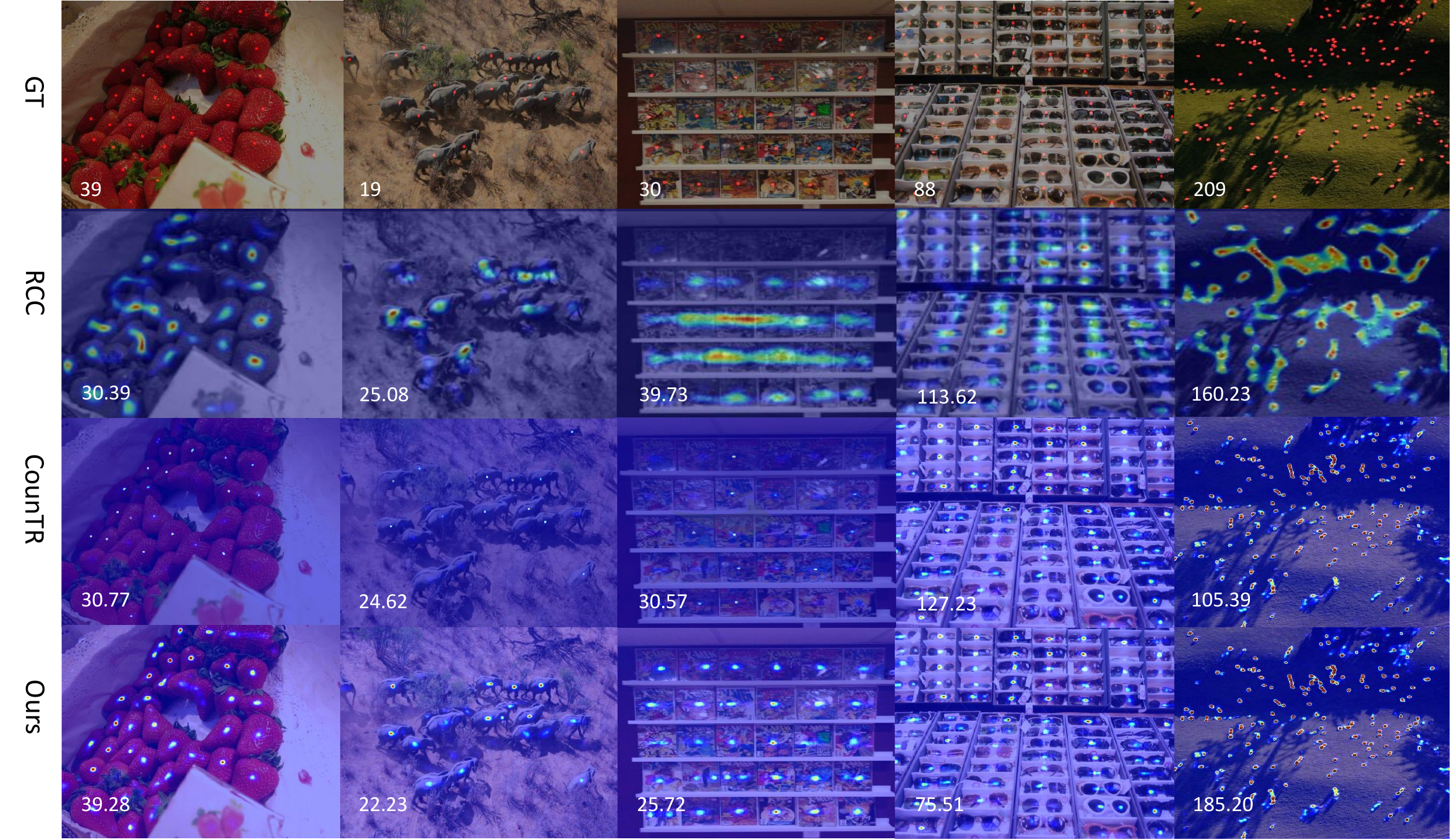}
\caption{Visual comparisons of the generated density maps with CounTR~\cite{liu2023countr} and RCC~\cite{hobley2022learning} on the FSC-147 dataset.}
\label{fig:compare1}
\end{figure}

We visually compare the density maps with CounTR~\cite{liu2023countr} and RCC~\cite{hobley2022learning} on the FSC-147 dataset. As shown in Fig.~\ref{fig:compare1}, our method can capture fine-grained details of objects. In contrast, CounTR sometimes generates density maps that do not accurately distinguish between individual objects in the map, \eg, in the fourth column, CounTR can't identify the far-way small fruit, while our method can.

\subsection{Cross-Domain Evaluation}
Following~\cite{wang2024vision}, we conduct a cross-domain evaluation, \ie, training the model on the FSC-147 dataset~\cite{ranjan2021learning} while directly evaluating on the CARPK dataset~\cite{hsieh2017drone}. The results are summarized in Table~\ref{table2}. The results for all the compared methods are obtained from~\cite{hobley2022learning,liu2023countr,djukic2023loca} under the same settings. As shown in Table~\ref{table2}, our model shows superior cross-domain performance compared with other methods, achieving a performance gain of 0.56 on MAE and 0.61 on RMSE compared to the previous best-performing method CounTR. Compared to the earlier EFC model~\cite{hobley2022learning}, the gains are even more significant, highlighting GCA-SUNet's superior generalization over the compared methods.

\begin{table}[htb]
\caption{Training on the FSC-147 dataset~\cite{ranjan2021learning} while testing on the CARPK dataset~\cite{hsieh2017drone}.} 
    \label{table2}
	\centering
    \renewcommand{\arraystretch}{0.8} 
	\setlength{\tabcolsep}{10pt}

	\begin{tabular}{clcc}
	\toprule
	\multirow{2}[2]{*}{Category} &\multirow{2}[2]{*}{Method} & \multicolumn{2}{c}{Test Set} \\
	\cmidrule(lr){3-4}
	& & MAE & RMSE \\
    \midrule
    \multirow{2}{*}{CAC}
    &LOCA~\cite{djukic2023loca}~\footnotesize{ICCV'23}   & 16.84 & 19.72 \\
    &CounTR~\cite{liu2023countr}~\footnotesize{BMVC'22}   & 11.52 & 14.56 \\
    \midrule
    \multirow{2}{*}{EFC}
    &RCC~\cite{hobley2022learning}~\footnotesize{CVPR'23} & 21.38 & 26.61 \\
    \cmidrule{2-4}
	& \textbf{Proposed GCA-SUNet} & \textbf{10.96} & \textbf{13.95} \\
\bottomrule
\end{tabular}

	\label{tab2}
\end{table}

\subsection{Ablation Study of Major Modules}
\label{sec:ablation_study}

We ablate the three major modules of proposed method on the FSC-147 dataset~\cite{ranjan2021learning}. Swin-UNet~\cite{swinunet} serves as the baseline model. 
The results are summarized in Table \ref{table:ablation}. The GCAM module alone significantly reduces the MAE by 2.32 on the test set and by 1.72 on the validation set compared with the baseline, underscoring its capability to enhance feature selectivity that is essential for complex scenes. 
Similarly, utilizing GEFS or GAFU individually also greatly reduces errors on both test and validation sets, showcasing the proposed gate mechanism's ability to highlight relevant features while suppressing irrelevant ones. The combination of these modules also achieves significant performance gains. Specifically, the overall framework with all three components achieves the most substantial improvement compared with the baseline, reducing the MAE by 4.78 on the test set and by 5.83 on the validation set. The results show the effectiveness of the synergistic interaction among different modules and emphasize the individual contribution of each major component.

\begin{table}[!t]
\caption{Ablation study of each component on the FSC-147 dataset~\cite{ranjan2021learning}.} 
    \label{table:ablation}
	\centering
    \renewcommand{\arraystretch}{0.8} %
	\setlength{\tabcolsep}{6pt}
	\begin{tabular}{cccccccc}
	\toprule
        \multirow{2}{*}{$\mathcal{F}^\textsc{GCAM}$} &
        \multirow{2}{*}{$\mathcal{F}^\textsc{GEFS}$} &
        \multirow{2}{*}{$\mathcal{F}^\textsc{GAFU}$} &
        \multicolumn{2}{c}{Test Set} & \multicolumn{2}{c}{Val Set} \\
 	\cmidrule(lr){4-5} \cmidrule(lr){6-7}
	 & & &MAE & RMSE & MAE & RMSE \\
	
    \midrule
    \XSolidBrush & \XSolidBrush & \XSolidBrush &  18.86 & 89.00    & 22.85& 85.16   \\
    \Checkmark & \XSolidBrush & \XSolidBrush & 16.54 & 85.20   & 21.13& 81.30     \\
    \XSolidBrush & \Checkmark &  \XSolidBrush &  16.63 & 80.33 &  20.13& 62.98     \\
    \XSolidBrush & \XSolidBrush & \Checkmark & 16.87& 84.36   &   21.60& 79.37   \\
    \Checkmark & \XSolidBrush &\Checkmark &  15.27& 84.81  &  20.08&81.58  \\
    \XSolidBrush &\Checkmark & \Checkmark & 15.07 & 78.79 & 18.58 & 59.17  \\
    \Checkmark & \Checkmark & \XSolidBrush & 15.21& 85.71  &  17.97& 55.67  \\
    \midrule
    \Checkmark & \Checkmark & \Checkmark & 14.08 & 84.77 & 17.02 & 55.37\\
 \bottomrule
	\end{tabular}
	\label{tab:abalation_factors}
\end{table} %


\subsection{Analysis of GCAM}
\begin{figure}[h]
\centering
\includegraphics[width=\linewidth]{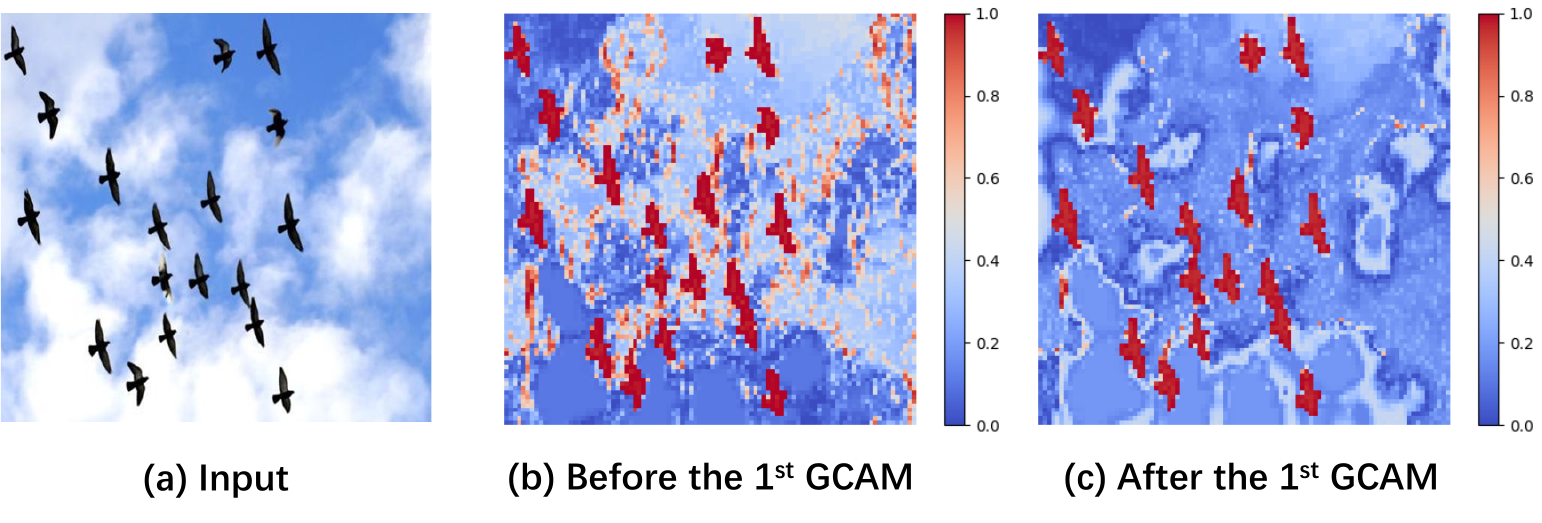}
\caption{Feature maps before/after the $\rm 1^{st}$ GCAM in the Swin-T enconder.}
\label{fig:compare2}
\end{figure}
As shown in Table~\ref{tab:abalation_factors}, GCAM contributes most significantly to the overall performance gain. We further visualize its effects as the density maps in Fig.~\ref{fig:compare2}. Following the practice in~\cite{hobley2022learning}, sub-figure (b) and (c) are obtained by projecting the density maps into two-dimensional spaces. Clearly by applying GCAM, the noisy estimation in the background areas (sky) is significantly reduced, with the foreground objects (birds) becoming more prominent. The results show that the proposed GCAM effectively enhances the representation of foreground tokens while suppressing irrelevant ones in the background.

\vspace{-4pt}
\begin{figure}[h]
    \centering
    \includegraphics[width=1\linewidth]{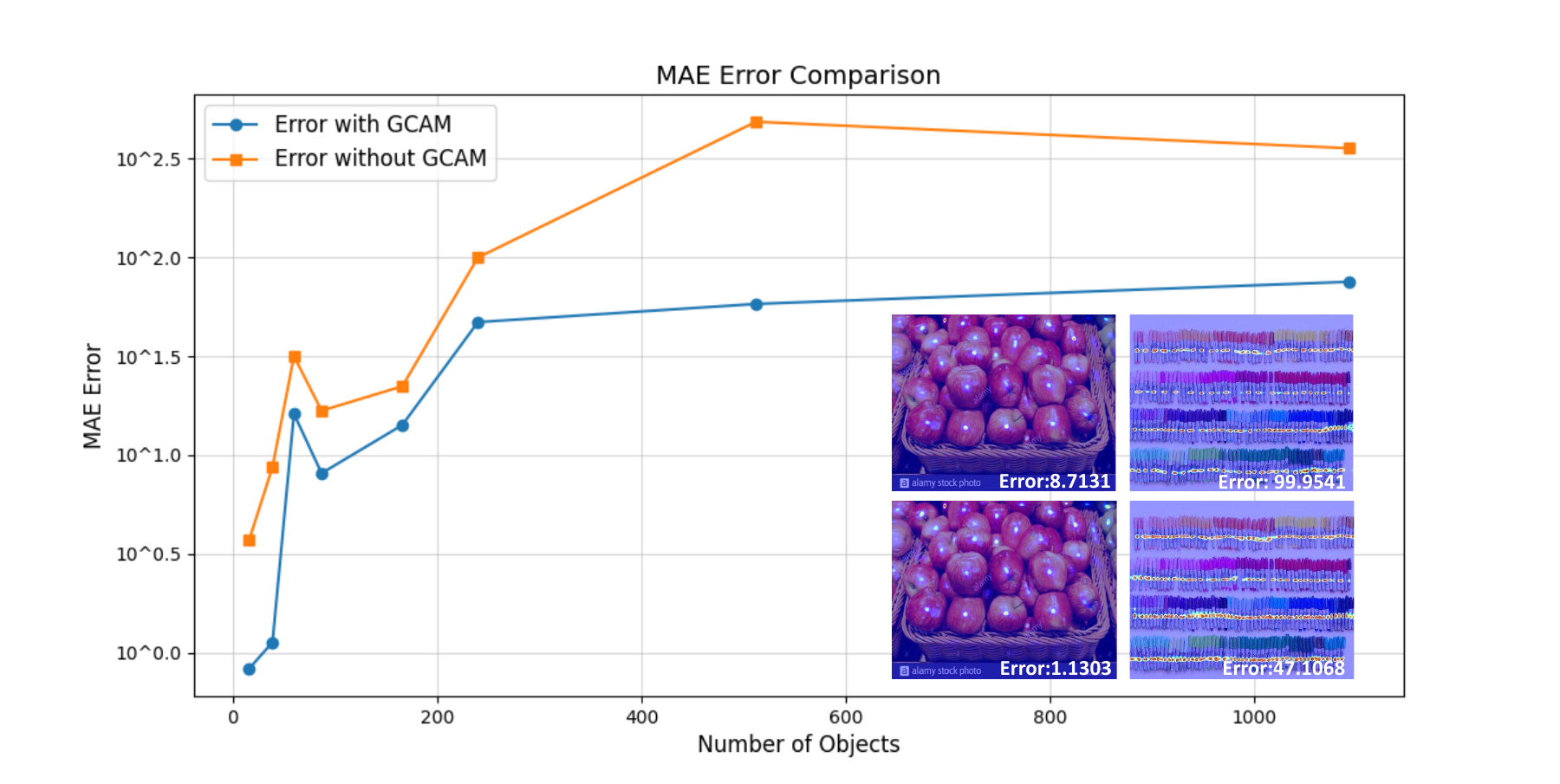}
    \vspace{-24pt}
    \caption{Evaluation of GCAM in various object density scenarios.}
    \label{fig:error_comparison}
\end{figure}

We further compare the results with/without the GCAM across scenarios of different object densities in Fig.~\ref{fig:error_comparison}.
For different object densities, the GCAM could consistently reduce the MAE, signifying its importance. On the other hand, Fig.~\ref{fig:error_comparison} reveals a clear trend of increasing error as the number of objects grows. Indeed, as the density of objects increases, they become smaller and counting tasks become more challenging. Future work will focus on optimizing the proposed approach to improve its performance in such challenging scenes.

\subsection{Model Complexity Analysis}

Tab.~\ref{tab:para-comp} compares the inference time, model size, FLOPs and training epochs with the previous best method, CounTR~\cite{liu2023countr}. 
The proposed model requires inference time comparable to CounTR. It significantly outperforms CounTR in terms of accuracy, at a higher but reasonable computational cost. 
 

\vspace{-6pt}
\begin{table}[htb]
\centering
\caption{Comparison of inference time, model size and FLOPs, and training epochs.}
\begin{tabular}{lcccc}
\toprule
    \multirow{2}{*}{Method} &
    \multicolumn{1}{c}{Inf. time} &
    \multicolumn{1}{c}{Parameters} &
    \multicolumn{1}{c}{FLOPs} &
    \multicolumn{1}{c}{Training} \\
        & 
    \textbf{(s)} & 
    \textbf{(M)} & 
    \textbf{(G)} & 
    \textbf{(epochs)} \\

\midrule
CounTR~\cite{liu2023countr} & 0.08 & 100  & 84  & 1000 \\
Proposed GCA-SUNet         & 0.11 & 192  & 173 & 300 \\
\bottomrule
\end{tabular}
\label{tab:para-comp}
\end{table}

\section{Conclusion}

The proposed GCA-SUNet effectively tackles the problems of exemplar-free counting by using a Swin-UNet architecture to directly map the input image to the density map of countable objects. The proposed GCAM exploits the attention information among the tokens of repetitive objects through the self-similarity matrix, and suppresses the features of irrelevant objects through a gate mechanism. The gate mechanism is also incorporated into the GEFS module and the GAFU module, which highlight the features most relevant to countable objects while suppressing irrelevant ones. Our experiments on the FSC-147 and CARPK datasets demonstrate that GCA-SUNet outperforms state-of-the-art methods, achieving superior performance in both intra-domain and cross-domain scenarios.

\bibliographystyle{IEEEtran}
\bibliography{icme2025references}


\end{document}